\documentclass{article}

    \usepackage[preprint]{neurips_2025}

\usepackage[utf8]{inputenc} % allow utf-8 input
\usepackage[T1]{fontenc}    % use 8-bit T1 fonts
\usepackage{hyperref}       % hyperlinks
\usepackage{url}            % simple URL typesetting
\usepackage{booktabs}       % professional-quality tables
\usepackage{amsfonts}       % blackboard math symbols
\usepackage{nicefrac}       % compact symbols for 1/2, etc.
\usepackage{microtype}      % microtypography
\usepackage{xcolor}         % colors

\usepackage{algorithm}
\usepackage{algpseudocode}
\usepackage{graphicx}
\usepackage{subcaption}

\title{Retrieval Augmented Decision-Making: A Requirements-Driven, Multi-Criteria Framework for Structured Decision Support}

\author{%
  Hongjia Wu \\
  State Key Lab of CAD\&CG \\
  Zhejiang University \\
  Hangzhou, China \\
  \texttt{hongjiawu@zju.edu.cn} \\
\And
Hongxin Zhang \\
  State Key Lab of CAD\&CG \\
  Zhejiang University \\
  Hangzhou, China \\
  \texttt{zhx@zju.edu.cn} \\
\And
Wei Chen \\
  State Key Lab of CAD\&CG \\
  Zhejiang University, \\
  Hangzhou, China \\
  \texttt{chenvis@zju.edu.cn} \\
\And
Jiazhi Xia \\
  School of Computer Science and Engineering \\   
  Central South University \\
  Changsha, Hunan, China \\
  \texttt{xiajiazhi@csu.edu.cn} \\
}

\begin{document}

\maketitle

\begin{abstract}

  Various industries have produced a large number of documents such as industrial plans, technical guidelines, and regulations that are structurally complex and content-wise fragmented. This poses significant challenges for experts and decision-makers in terms of retrieval and understanding. Although existing LLM-based Retrieval-Augmented Generation methods can provide context-related suggestions, they lack quantitative weighting and traceable reasoning paths, making it difficult to offer multi-level and transparent decision support. To address this issue, this paper proposes the RAD method, which integrates Multi-Criteria Decision Making with the semantic understanding capabilities of LLMs. The method automatically extracts key criteria from industry documents, builds a weighted hierarchical decision model, and generates structured reports under model guidance. The RAD framework introduces explicit weight assignment and reasoning chains in decision generation to ensure accuracy, completeness, and traceability. Experiments show that in various decision-making tasks, the decision reports generated by RAD significantly outperform existing methods in terms of detail, rationality, and structure, demonstrating its application value and potential in complex decision support scenarios.
  
\end{abstract}

\section{Introduction}

With the continuous advancement of information technology, various industries have generated a large volume of diverse text materials, including industrial plans, technical guidelines, and regulations [1,2]. These documents contain rich domain knowledge and decision-making references, but their dispersed content and complex structure pose significant challenges for experts and decision-makers in terms of retrieval and understanding [3,4]. How to efficiently extract [5], organize, and present key information from massive batches of text to support downstream decision-making and review [6] has become an urgent issue for both academia and industry.

Large language models (LLMs) have shown excellent capabilities in understanding and generating human-like language, and they have been widely applied to various complex reasoning and decision-making tasks [7–9]. In recent years, researchers have begun exploring their potential in decision support, mainly by retrieving relevant information from large-scale text corpora and generating outputs in natural language form [8,9]. However, most existing methods rely on the Retrieval-Augmented Generation (RAG) mechanism. Although RAG can provide context-related information, it often lacks a traceable [10,11] and quantifiable computation process [12], making it difficult to build hierarchical decision support structures. This limits the ability to find optimal solutions in specific decision-making scenarios.

In key areas such as industrial planning, technology roadmap development, and regulatory compliance, decision-making requires not only semantic understanding of heterogeneous texts [13,14], but also a transparent analysis process that combines qualitative insights with quantitative trade-offs [15]. Traditional decision analysis methods, such as the Analytic Hierarchy Process (AHP) [16], have been widely used for such tasks. However, these methods usually depend on expert knowledge and extensive manual effort, making them difficult to scale efficiently.

To address this, we propose a hybrid method that integrates multi-criteria decision analysis with the semantic understanding of language models—Retrieval Augmented Decision-Making (RAD). This method can automatically construct a weighted hierarchical decision model from batches of industry documents and generate decision reports based on it. Unlike existing LLM-based approaches that generate suggestions only from document passages, our framework introduces explicit, weighted reasoning paths in the decision-making process to ensure the accuracy, completeness, and traceability of the results. Experimental results show that this method significantly outperforms existing approaches in generating more detailed, reasonable, and well-structured decision reports.

The main contributions of this paper are as follows:

1. A hierarchical decision criteria modeling method is proposed, addressing the limitations of existing decision-generation approaches in semantic depth and structured representation.

2. A multi-agent mechanism for assigning weights to decision criteria is designed, enabling automated and interpretable quantification of semantic units at different levels, thus providing more detailed support for decision-making.

3. An end-to-end decision generation process is developed, which produces accurate, complete, and traceable decision results while maintaining high-quality natural language expression.

\section{Background}

\subsection{Multi-Criteria Decision-Making Methods}

The decision-making process is often viewed as an input–process–output system: first, collecting multi-source information (such as data, expert evaluations, stakeholder preferences, and social norms); then, conducting quantitative and qualitative analysis during the "process" stage; and finally, producing executable decision results (such as plan rankings, recommended options, or implementation paths). The AHP and Interpretive Structural Modeling (ISM) [17,18] are commonly used to describe the hierarchy and causal relationships among elements. For example, [19] applies these methods to identify barriers to IoT implementation in manufacturing, and Lei et al. [20] use them to evaluate urban environmental quality. Although these methods offer logical rigor in quantitative analysis, they rely heavily on manual construction.

To improve modeling efficiency, researchers have explored combining structured frameworks with text mining techniques [21,22]. For example, Zhang et al. [23] applied this approach to analyze the similarity of regional policy texts. In addition, methods based on causal graphs or semantic networks [24,25] have been introduced to semi-automatically construct execution and feedback paths. Drury et al. [26] reviewed causal relation extraction techniques and discussed the challenges of semantic uncertainty and consistency in structured modeling.

LLMs have recently been used as general decision support tools across many fields. Researchers have explored methods to integrate LLMs into decision workflows [27,28], such as using chain-of-thought prompts to trigger step-by-step reasoning or employing RAG to incorporate domain knowledge [7,8]. However, existing methods lack traceable and quantifiable computation mechanisms [10–12]. Recent large-model tools, such as Deep Research, can break user queries into multiple subtasks, perform multi-round internet searches and data analysis, and finally synthesize detailed reports, which can also be used for decision reference.

Our method differs by automatically constructing a clear, weighted hierarchical decision model during the decision generation process. It organically integrates Multi-Criteria Decision Making with the semantic understanding ability of LLMs, ensuring the accuracy, completeness, and traceability of the decision results.

\subsection{Text Structure Parsing}

Text structuring and parsing aim to extract structured information such as entities, relationships, and events from unstructured text and organize it into a form that is easy for computers to process [29]. Early methods mainly relied on patterns and rules defined by domain experts and have been widely applied in financial analysis [30,31], legal, and historical analysis [32,33]. However, these methods have high costs for rule maintenance and cross-domain transfer.

Research based on LLMs has surged recently. Generative information extraction uses prompt engineering to directly generate structured label sequences from the model [34], eliminating complex post-processing and greatly improving extraction efficiency and generalization. For example, Liu et al. proposed the SciIE framework, which uses GPT-4 with multi-round prompting to automatically build domain knowledge graphs from scientific papers [35]. Methods like RAG and its extension GraphRAG retrieve external text segments and combine graph-structured entity–relationship information to automatically parse unstructured text into structured representations such as entities and relationships, supporting precise execution of downstream tasks [9,36]. Multi-agent architectures have been introduced for text parsing tasks, where different agents handle subtasks like entity recognition, relation extraction, and context verification, and their results are combined into a unified structured output [37,38].

The application of large models in text structuring has provided important insights for our research. Unlike existing methods, our approach uses document outlines to segment semantic text blocks and performs targeted information extraction based on user needs. For information integration, we introduce a multi-agent architecture to simulate manual evaluation processes, achieving more efficient and semantically consistent data processing.

\section{Methods}

\subsection{Decision Making Process}

The decision-making process must consider both quantitative information (such as statistics and metric measurements) and qualitative factors (such as value judgments and social norms), while also handling heterogeneous preferences from multiple stakeholders to produce actionable decisions. In this framework, inputs reflect the environment and stakeholder needs, and outputs are practical plans or conclusions, both of which critically affect decision quality and subsequent implementation. The "process" stage involves information processing and evaluation.

Following this process framework, we define the user’s decision request as $x=\{T,d,S\}$, where $T$ represents reference documents for decision support; $d$ is the description of the decision problem; and $S=\{s_i\}_{i=1}^{o}$ denotes $o$ candidate options to be evaluated. The output $y$ includes decision criteria, option evaluation scores, reasoning for judgments, and decision recommendations.

\subsection{RAD Overview}

\begin{figure}
    \centering
    \includegraphics[width=1\linewidth]{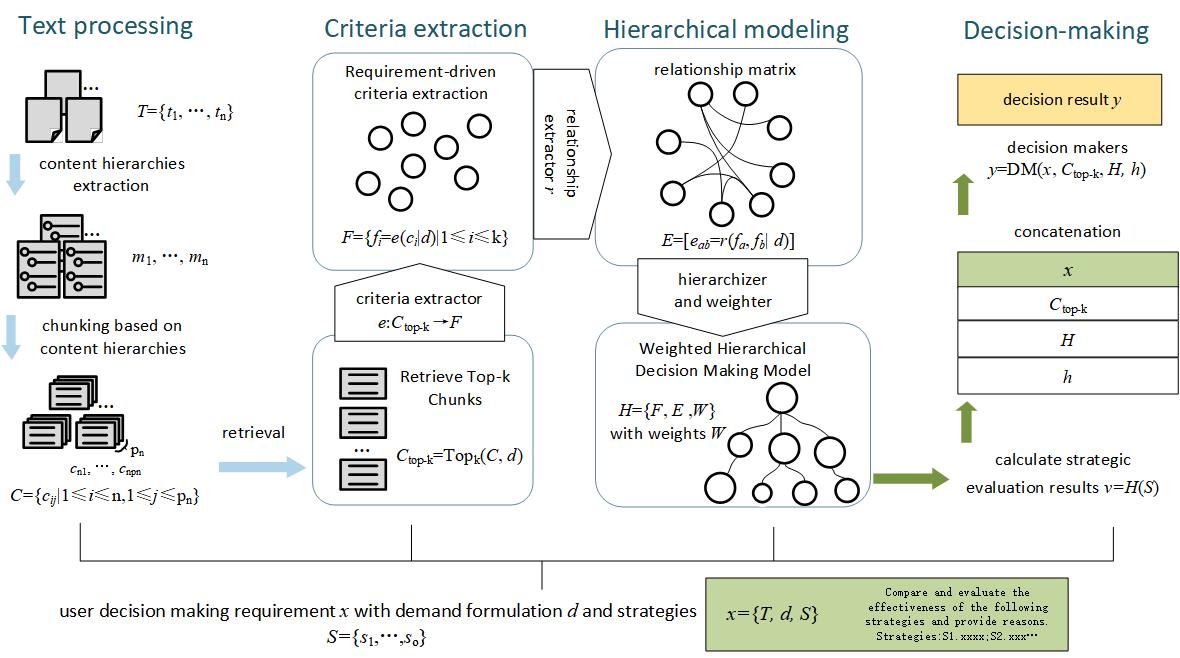}
    \caption{The pipeline of the method.}
    \label{fig:1}
\end{figure}

We propose RAD, as shown in Fig. \ref{fig:1}. Given a decision request $x$, the method generates and outputs the decision result $y$. As illustrated in Fig. 1, the process first retrieves and extracts a set of key criteria $F$ based on $x$. Next, it constructs a weighted hierarchical tree model $H$ using $x$ and $F$. Then, $H$ is used to quantitatively evaluate the candidate options in $x$, producing an intermediate result $v$. Finally, by integrating $x$, $F$, $H$, and $v$, the decision result $y$ is output. Each step will be described in detail in this chapter.

\subsection{Text Processing}

To obtain decision criteria, the document set $T=\{t_i\}_{i=1}^{n}$ must be processed. Through the document processing workflow, documents are divided into text blocks to allow large models to extract decision criteria from each block. Research [39] shows that using the document’s table of contents structure effectively supports subsequent text analysis tasks. The document segmentation method based on hierarchical directory structure (by identifying semantic markers like chapter titles) preserves the semantic integrity and contextual relevance of the text. For criteria extraction, this method not only focuses on the core features of decision criteria but also filters out redundant details, optimizing the density of decision information. The document processing workflow includes a multi-level directory extraction algorithm and an adaptive text chunking strategy.

\subsubsection{Content Hierarchies Extraction}

When documents have hierarchical headings, parsing the table of contents directly provides natural segmentation points, with support for up to three levels. For documents without titled structures, the table of contents is generated by dividing and encoding each paragraph, then clustering to determine parent levels, forming a three-level directory structure. Finally, a large model generates clear and concise titles for each level. This results in a hierarchical directory $\{m_1,...,m_n\}$.

\subsubsection{Text Segmentation}

Using string matching, the first sentence of each directory entry is matched to determine the final text block boundaries. Ultimately, each text file is divided into $p_n$ text blocks, forming the set $C=\{c_{ij}|1\leq i\leq n,1\leq j\leq p_n\}$. These text blocks are encoded and stored in a vector database, providing a foundation for subsequent criteria extraction. The text block embedding method can use Late Chunking [40].

\subsection{Criteria Extraction}

The goal of criteria extraction is to identify key criteria from the text that influence decisions or outcomes. These criteria may be abstract, causal, or implicit. The focus is on the underlying logical relationships and causal chains in the text, rather than specific entities. In the supplementary material, we provide cases illustrating the differences between this method and entity recognition.

In the criteria extraction stage, the top-k most relevant text blocks $C_{top-k}=Top_k(C,d)$ are retrieved from the vector database based on user demand $d$. The text blocks $C_{top-k}$ are then input into the criteria extractor $e:C_{top-k}\rightarrow F$, resulting in $k$ criteria $F=\{e(c_{i}|d)|1\leq i\leq k\}$.

We use an LLM as the criteria extractor, which understands the semantics of text segments and outputs summarized criteria terms along with the relevance between each segment and the user’s request. To support further analysis, structured key criteria such as time and quantity can also be extracted.

Although the document processing stage provides section titles that partly summarize the content of each text block, we conduct a second round of key information extraction to better match the user’s request. This is based on two considerations: first, different documents may have identical titles but very different content; second, section titles often offer only broad summaries, which may not accurately reflect the specific aspects relevant to the user’s needs.

\subsection{Hierarchical Modeling}

The hierarchical modeling process structures complex decision problems, making it easier to compare and quantify the importance of each criterion step by step. This improves the logic, transparency, and accuracy of decision-making. The process includes criterion relationship extraction, criterion layering, and factor weighting.

\subsubsection{Relationship Extraction}

The relationship extractor $r(f_a,f_b|d)$ outputs $e_{ab} \in \{0,1\}$ to determine whether there is a direct influence or causal relationship from $f_a$ to $f_b$ under the given user requirement $d$, where $f_a, f_b \in F$. Pairwise comparisons are performed for all criteria in the set $F$, resulting in an association matrix $E=[e_{ab}]_{k\times k}$, which serves as the basis for subsequent criterion layering.

\subsubsection{Criteria Hierarchization}

We adopt the ISM method for the criterion hierarchization process, using $E$ to divide all criteria $F$ into multiple levels. Higher-level criteria usually represent final outcomes or goals, while lower-level criteria tend to serve as causes or driving factors. This provides a clear logical foundation for subsequent quantitative weighting.

\subsubsection{Criteria Weighting}

The weighting process uses a multi-agent voting mechanism to simulate how domain experts score during decision-making. By inputting the list of criteria at each level, the system outputs the average importance ranking from all agents. Based on these rankings, we apply AHP to assign corresponding weights to the key criteria.

In this process, we define $A_{M}$ as the manager, who assigns tasks to subordinate agents and integrates their results to construct a weight distribution based on the hierarchical model. $A_M$ supervises five domain experts, $A_{e1}$ to $A_{e5}$. Based on the decision task $d$, $A_M$ identifies five relevant fields and assigns a role to each expert accordingly. Each expert then ranks the importance of criteria at the same level from their domain perspective and provides reasons for their rankings. $A_M$ evaluates the validity of these reasons, integrates the rankings, and calculates the average order. Then, using the AHP method, the average ranking is converted into specific weights for each key criterion, resulting in the final weight vector $W = [w_1, ..., w_k]$.

Finally, the hierarchical criteria $F$, the relation matrix $E$, and the weights $W$ are combined to form the Weighted Hierarchical Decision Making Model $H = \{F, E, W\}$.

\subsection{Decision-making}

By combining the structured decision-making framework of the AHP model with the natural language processing capabilities of large language models, the decision-making process enables both quantitative evaluation and qualitative analysis of text-based alternatives.

For the set of alternatives $S$ provided by the user, the large language model scores each alternative $s_i$ against each criterion $f_j$, using the AHP 1–9 scale. This results in a score matrix $A = [a_{ij}]_{o \times k},\ where\ a_{ij} \in [1, 9]$. A consistency check is then performed on $A$ to ensure logical coherence in the scoring. Based on the score matrix $A$ and the weight vector $W$, the overall score of each alternative is calculated as $V_i = \sum_{j=1}^k w_j \cdot a_{ij}, \quad i = 1, 2, \dots, o$. Alternatives are then ranked according to their overall scores $v_i$ to determine priority.

In the decision generation stage, we combine $x$, $F$, $H$, and $v$ and input them into the large language model. The model analyzes each alternative and outputs its strengths and weaknesses under each criterion, along with an overall evaluation.

\section{Analysis}

The manual decision-making process usually involves two aspects: first, developing decision plans and evaluating their pros and cons by reviewing documents; second, comparing multiple alternatives to select the best option. Our experiment aims to verify the effectiveness of the proposed method in these two typical decision stages.

\subsection{Experiment 1}

This experiment guides the model to formulate plans by inputting relevant documents and decision problems. It compares the performance of the proposed method with its ablated versions in decision generation and further analyzes the differences with the graphRAG method.

\subsubsection{Dataset}

\textbf{R1.} Prompt Construction Dataset: This includes three materials related to ChatGPT prompt design, used to guide the creation of high-quality generation prompts. Among them, \textit{Reasoning Best Practices} and \textit{Text Generation and Prompting} come from the OpenAI developer platform, while \textit{The Art of Asking ChatGPT for High-Quality Answers} is authored by Ibrahim John. The dataset contains a total of 17,811 tokens and is divided into 47 text segments after document processing.

\textbf{R2.} COVID-19 Treatment and Management Guidelines: This includes official documents from the World Health Organization website, such as \textit{Clinical Management of COVID-19} and \textit{Therapeutics and COVID-19}, totaling 186,070 tokens. After document processing, it is divided into 200 text segments.

\subsubsection{Configuration}

We generated 50 decision requests for each dataset and produced decision recommendations using the following methods. Each request was generated five times, and the average was taken. Since there are no standard answers for decision generation tasks, we followed research [9] and used large language models to perform comparative evaluations of the generated results.

Comparison methods include:

\textbf{RAD.} The complete method proposed in this paper;

\textbf{NS.} The performance of our method using fixed-word segmentation;

\textbf{NH.} The performance of our method without input the hierarchical decision model, relying only on the extracted decision factors for reasoning;

\textbf{GR.} GraphRAG method with global search.

Under all comparison conditions, the context window size and the prompts used for answer generation remain consistent, with only slight adjustments in the style reference section to match the information needs of different context types. The deepseek-v3 model is used for generation. Evaluation criteria include:

\textbf{C1.} User need consistency: This metric measures the extent to which the generated answers respond to the user's actual needs and context, checking whether irrelevant or unsuitable suggestions are produced by ignoring the user context [41].

\textbf{C2.} Evidence-based decision reasoning: This metric assesses whether the answers provide verifiable and traceable evidence support, including quantitative methods such as data, models, algorithms, or historical results, avoiding subjective assumptions or unsupported statements [42].

\textbf{C3.} Logical consistency of reasoning: This metric examines whether the reasoning process is clear and coherent, maintaining causal relationships at each step to ensure the final conclusions are built on a solid chain of reasoning [43].

\subsection{Experiment 2}

This experiment aims to evaluate the performance of the proposed method in multi-option comparison tasks and to determine whether it can accurately identify the better options.

\subsubsection{Dataset}

The document\textit{Best Practices for Prompt Engineering with the OpenAI API} from the OpenAI official website, which is not included in R1, is used as the validation dataset for this experiment. This dataset contains seven pairs of positive and negative prompt examples. We use it to verify whether the model has effective capability for solution discrimination and comparison.

\subsubsection{Configuration}

This experiment uses the deepseek-v3 model for evaluation, with the following specific criteria:

\textbf{C4.} Whether the model can accurately distinguish positive (1) and negative (0) examples.

\textbf{C5.} Whether the constructed hierarchical decision model covers the key issues involved in the examples (1 for yes, 0 for no).

\textbf{C6.} Whether the positive examples score significantly higher than the negative examples on relevant factors within the hierarchical decision model (1 for yes, 0 for no).

\section{Result}

\subsection{Experiment 1}

\begin{figure}[H]
  \centering
  \begin{subfigure}[b]{0.3\textwidth}
    \includegraphics[width=\textwidth]{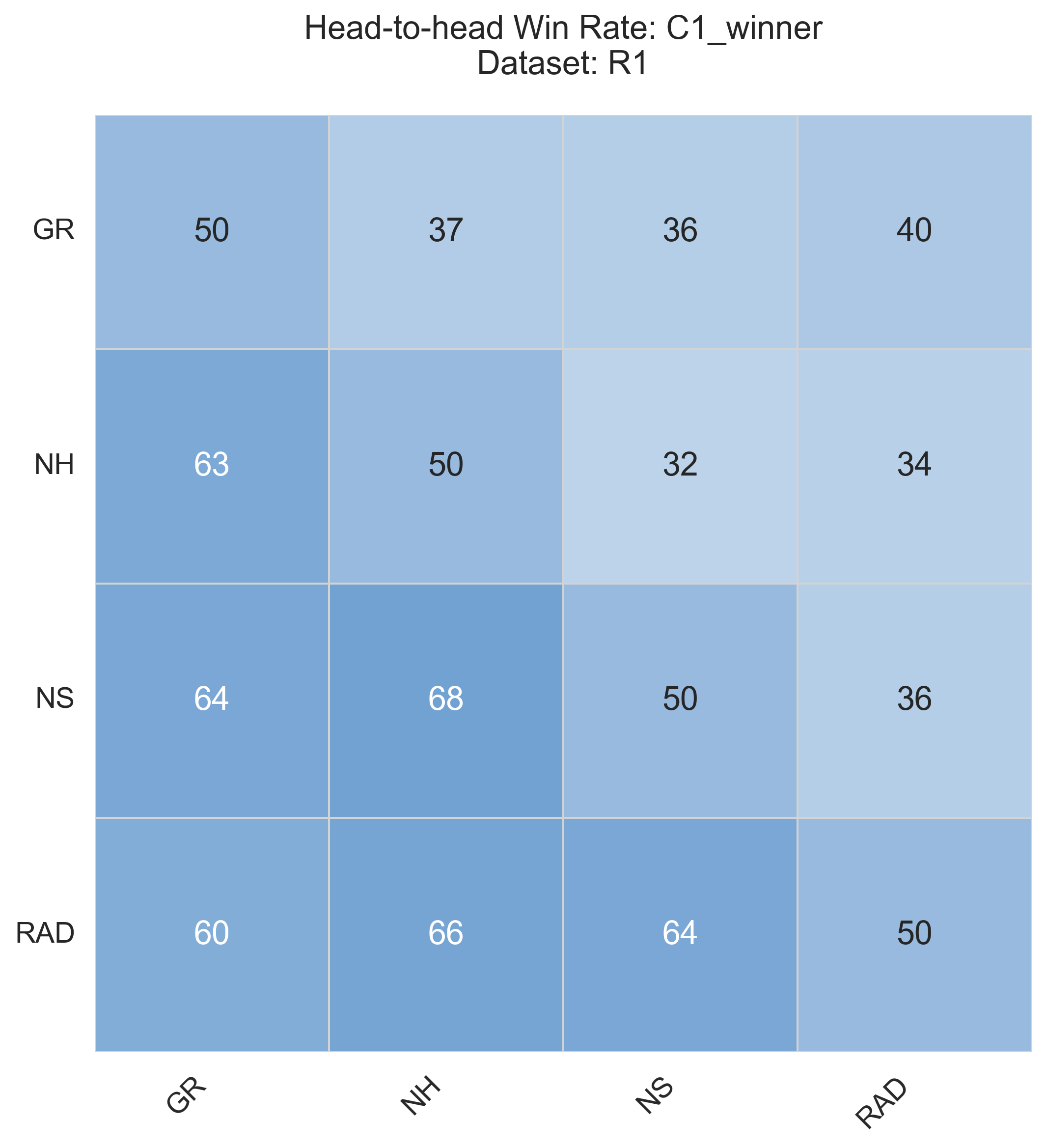}
    \label{fig:image1}
  \end{subfigure}
  \hfill
  \begin{subfigure}[b]{0.3\textwidth}
    \includegraphics[width=\textwidth]{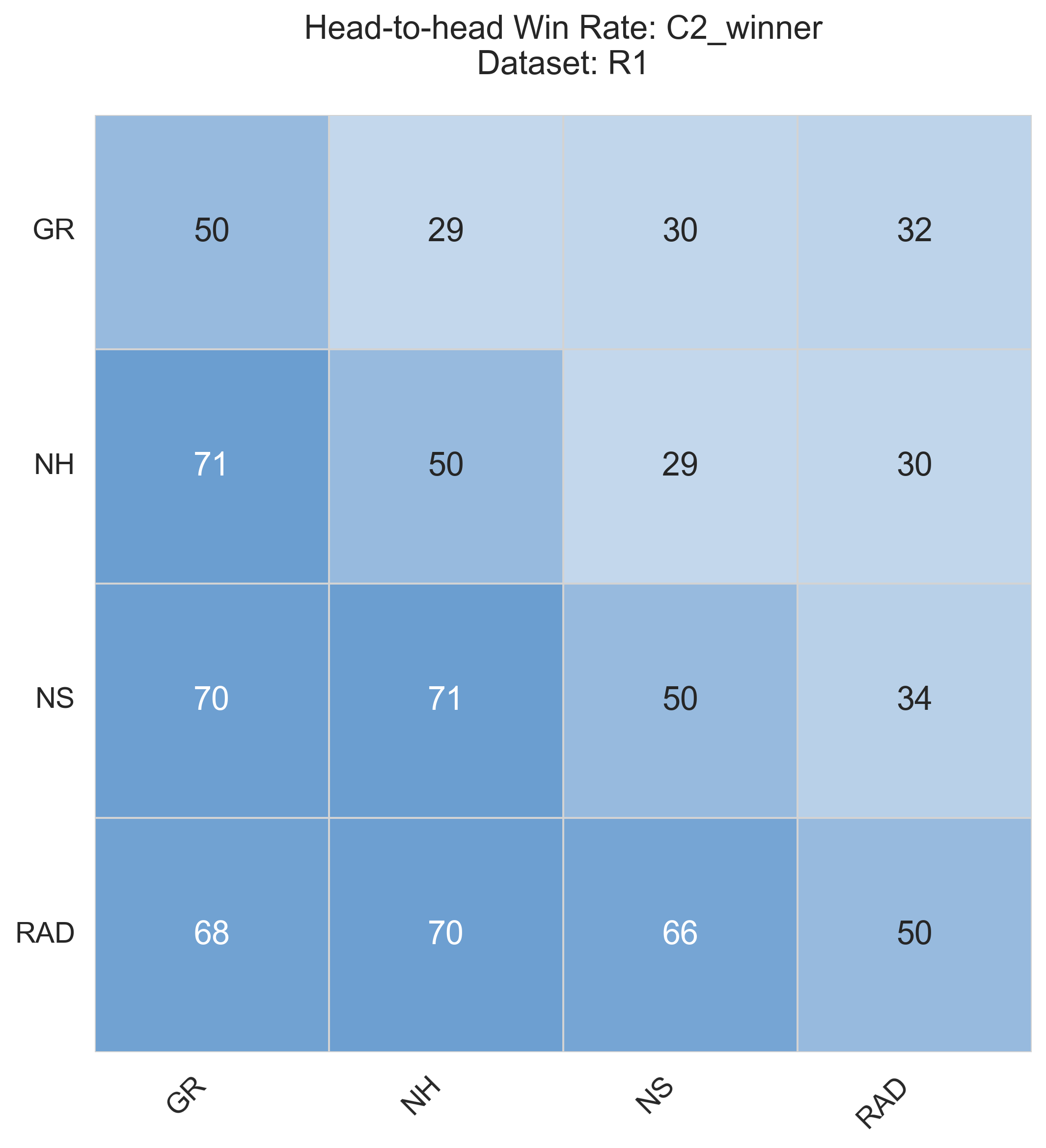}
    \label{fig:image2}
  \end{subfigure}
  \hfill
  \begin{subfigure}[b]{0.3\textwidth}
    \includegraphics[width=\textwidth]{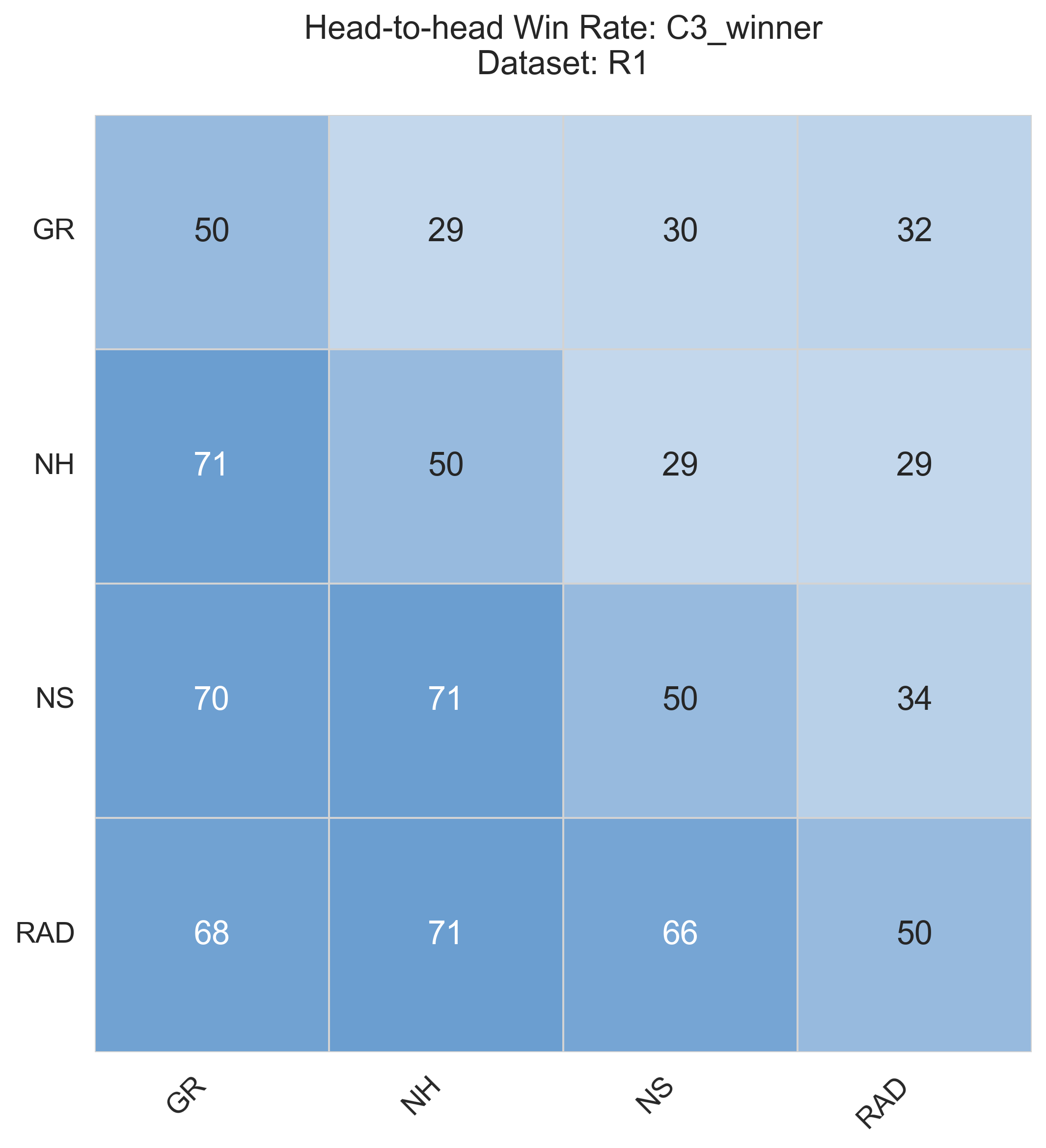}
    \label{fig:image3}
  \end{subfigure}
  \vspace{1em}
  \begin{subfigure}[b]{0.3\textwidth}
    \includegraphics[width=\textwidth]{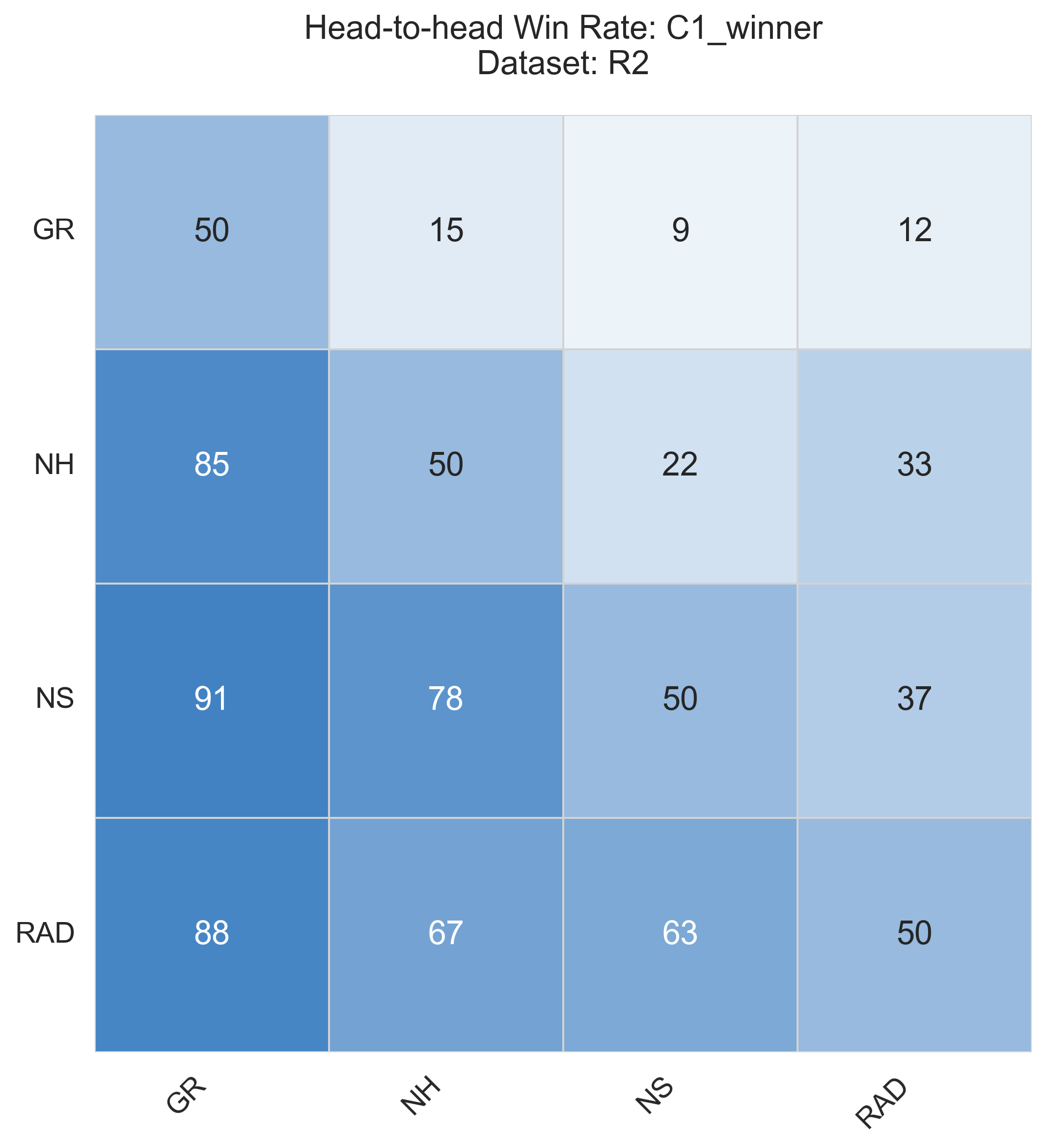}
    \label{fig:image4}
  \end{subfigure}
  \hfill
  \begin{subfigure}[b]{0.3\textwidth}
    \includegraphics[width=\textwidth]{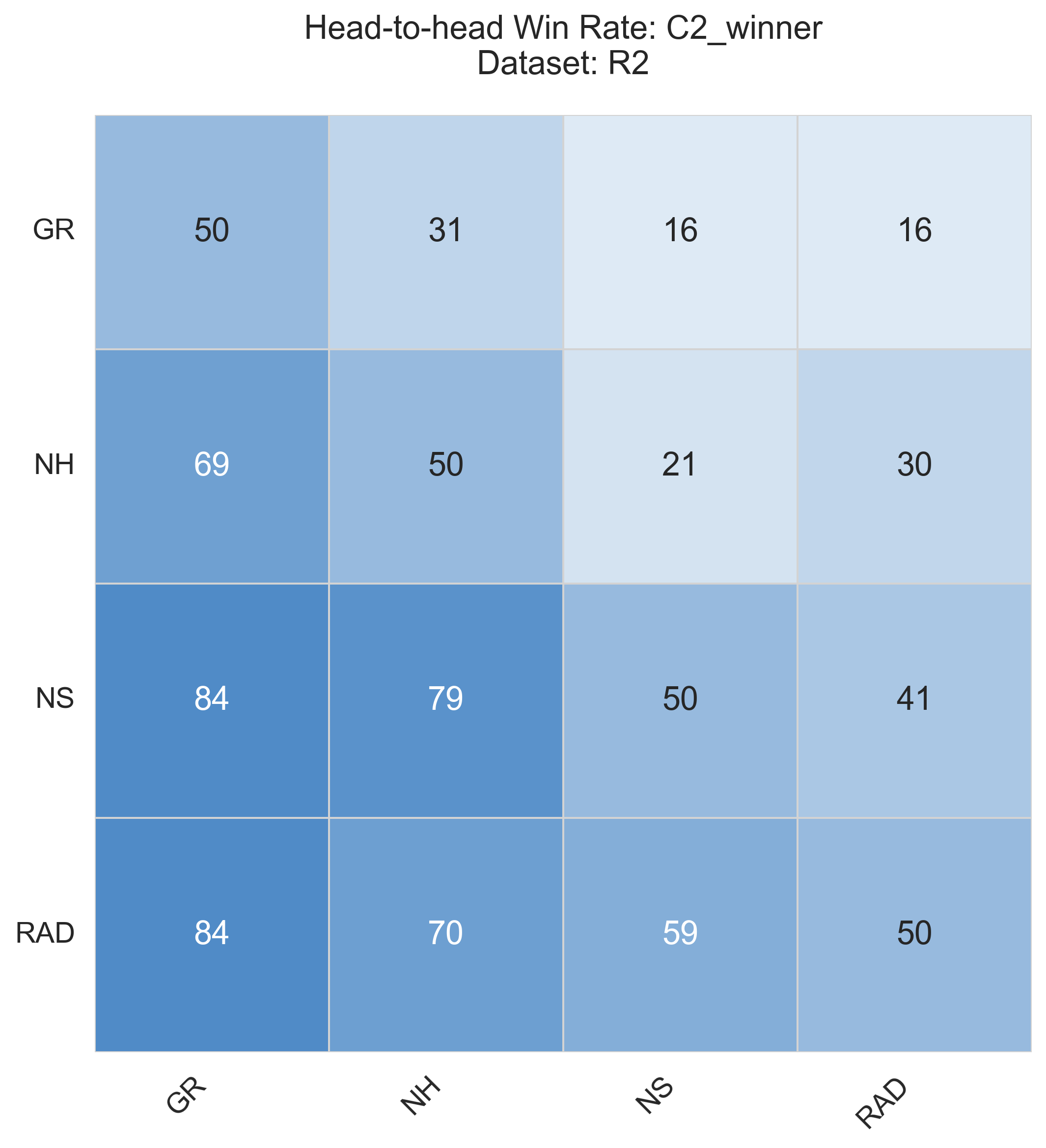}
    \label{fig:image5}
  \end{subfigure}
  \hfill
  \begin{subfigure}[b]{0.3\textwidth}
    \includegraphics[width=\textwidth]{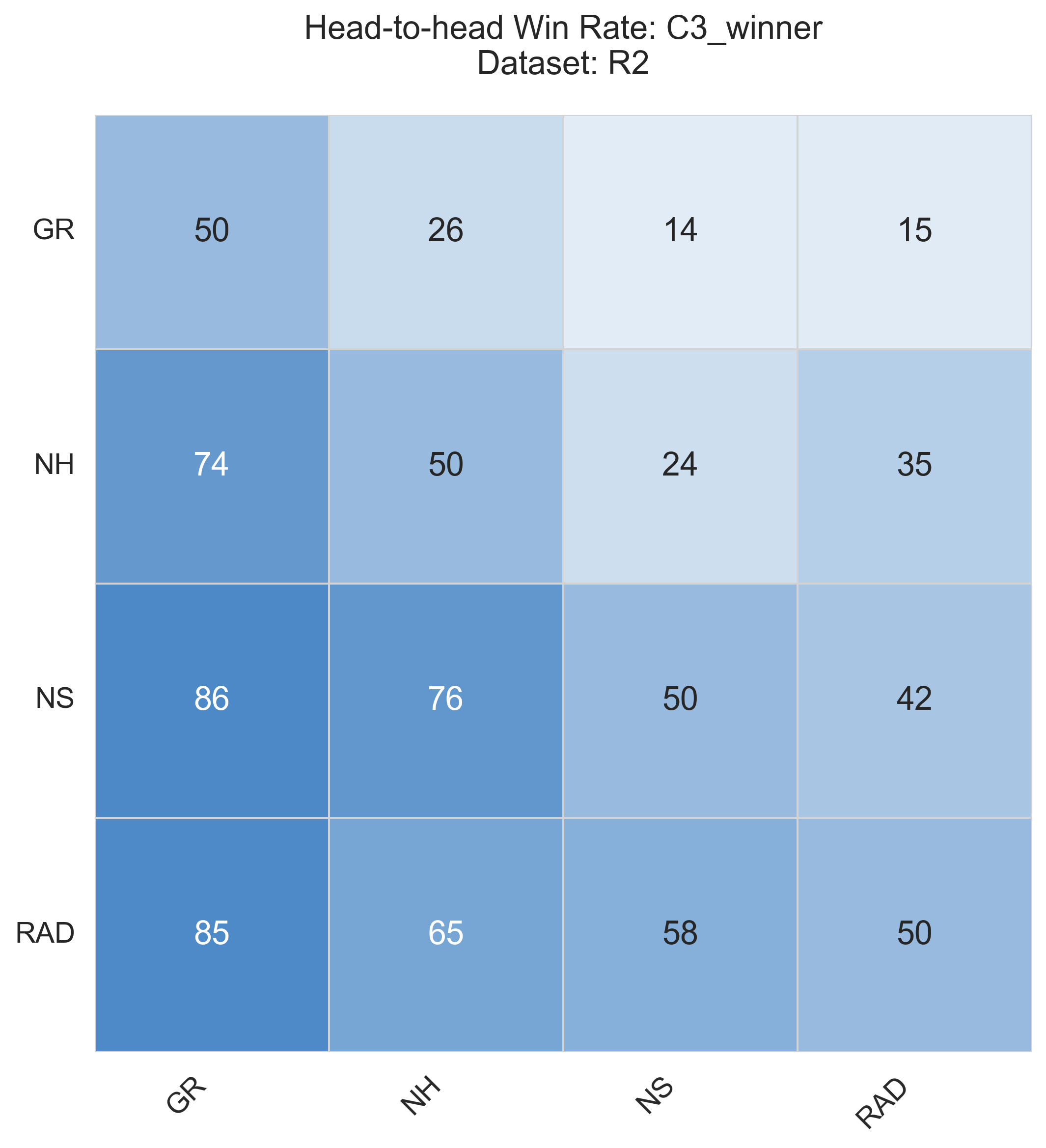}
    \label{fig:image6}
  \end{subfigure}

  \caption{The figure displays the direct win rate percentages of the row condition over the column condition across two datasets and four evaluation metrics, based on 50 questions per comparison. Self-comparison win rates (diagonal) are shown as the expected 50\% reference value.}
  \label{fig:wholefigure}
\end{figure}

The comparison results of Experiment 1 are shown in Fig. \ref{fig:wholefigure}.

\subsubsection{Comparative Analysis Between RAD and Ablation Versions}

When comparing RAD with NS, the experimental results show that RAD significantly outperforms NS on all metrics (p < 0.05). Especially in terms of user need consistency, RAD achieved a 63\% win rate in R2 (p = 0.0579), indicating that directory-based segmentation can effectively improve the alignment between generated content and actual user needs. This advantage may stem from the directory structure preserving the hierarchical relationships in the text, thus better capturing implicit requirements in the task. In R1, RAD’s win rate remained stable between 64\% and 66\%. Since the text token count in R1 is smaller, the complete hierarchical structure better captures the core criteria of the needs.

Further analysis comparing RAD and NH shows that the hierarchical decision model has a broader impact on generation quality. For C2, RAD achieves a win rate of 70\% in R1 (p < 0.0001), significantly outperforming NH, indicating that hierarchical decision-making effectively enhances the verifiability of generated answers. In addition, for C3, RAD maintains a win rate between 65\% and 71\% across both tasks, suggesting that this method better preserves the integrity of the reasoning chain and helps avoid logical gaps or subjective assumptions.

\subsubsection{Comparative Analysis Between RAD and GraphRAG}

Compared with GR, RAD shows overall advantages. In R2, GR's win rate on all metrics is below 15\%, with especially poor performance on C1. An analysis of GR’s outputs shows that its strategies are less focused on user needs, while RAD’s responses are more specific and aligned with detailed factors. In R1, GR performs slightly better, reaching a 40\% win rate on C1, but still lower than RAD’s 60\% (p = 0.0016).

Moreover, for C2 and C3, RAD maintains a stable win rate of around 68\%, further confirming its advantage in generating quantitatively supported and logically consistent responses. We observed that GR can generate numerical scores for options due to the embedded knowledge in the large model's parameters. However, most of these scores are general and cannot be traced to specific decision factors. Overall, GR has clear limitations, whereas RAD significantly improves the relevance and reliability of generated content by integrating document structure-based segmentation and a hierarchical decision model.

\subsection{Experiment 2}

We traced the decision-making computation process of the method and summarized the results of Experiment 2, as shown in Table 1.

\begin{table}
    \caption{The results of Experiment 2.}
    \centering
    \label{tab:ex2}
    \begin{tabular}{cccc|cccc}
    \toprule
         & C4 & C5 & C6 &  & C4 & C5 & C6 \\
         \midrule
        Case 1 & 0 & 0 & 0 & Case 2 & 1 & 1 & 1\\
        Case 3 & 1 & 1 & 1 & Case 4 & 1 & 1 & 1 \\
        Case 5 & 1 & 1 & 1 & Case 6 & 1 & 0 & 0 \\
        Case 7 & 0 & 0 & 0 &   &   &   &   \\
        \bottomrule
    \end{tabular}
\end{table}

As shown in Table \ref{tab:ex2}, the method's answers fully satisfy criteria C4, C5, and C6 in Cases 2, 3, 4, and 5. This is because the reference documents from R1 cover the key points required for these cases, such as providing examples in prompts and specifying the output format. However, misjudgments occurred in Cases 1, 6, and 7. By tracing the calculation and judgment process of RAD, we found that the R1 documents did not include the key decision points mentioned in these cases: "use punctuation mark to separate the instruction and context," "say what to do," and "use leading words." Nevertheless, Case 6 was still judged correctly under criterion C4 because the positive example scored higher on the criterion “the problem should be clearly described,” which allowed for correct distinction.

Therefore, through the answer generation process of RAD, we can clearly analyze the reasons behind each correct or incorrect judgment.

\section{Discussion}
The experimental results demonstrate that our proposed RAD framework significantly outperforms both ablation versions and baseline methods across multiple evaluation metrics and task domains. However, several limitations warrant discussion in the context of real-world decision support applications.

Although RAD demonstrates excellent performance in generating decision suggestions, its overall effectiveness is still highly dependent on the quality of the input documents. Since source materials from different industries often vary in terminology and detail level [1,2], this can affect the construction of the hierarchical decision model and the reasoning process. Moreover, when the source documents lack clear decision guidance, RAD may struggle to provide actionable plans, timelines, or resource allocation suggestions in its output. In such cases, general prompts may not be sufficient to meet users’ specific needs, and it may be necessary to design more tailored prompts for particular scenarios or fine-tune the model to better adapt to industry-specific terms and in-depth content.

In addition, the current RAD framework still lacks the ability to generate and integrate multimodal decision elements such as charts and flow diagrams, which also play an important role in decision-making [44,45]. Future work may consider building cross-modal hierarchical decision models to further expand the application of RAD in digital decision support systems.

\section{Conclusion}

We propose RAD, an innovative decision-generation framework that deeply integrates hierarchical decision modeling with the semantic understanding capabilities of LLMs. It aims to automatically generate structured, traceable, and evidence-based decision reports from documents that are structurally complex and semantically diverse. Architecturally, RAD introduces a hierarchical criterion evaluation mechanism, which improves the interpretability and transparency of the decision-making process. Meanwhile, the natural language understanding ability of LLMs enhances the accuracy of information extraction and reasoning. Experimental results show that RAD can generate decision suggestions on real-world data while maintaining logical consistency and offering a traceable reasoning path throughout the generation process.

\section*{References}

{
\small

[1] Grangel-González, I., \& Vidal, M. E. (2021, April). Analyzing a knowledge graph of industry 4.0 standards. In Companion Proceedings of the Web Conference 2021 (pp. 16-25).

[2] Redwood, J., Thelning, S., Elmualim, A., \& Pullen, S. (2017). The proliferation of ICT and digital technology systems and their influence on the dynamic capabilities of construction firms. Procedia Engineering, 180, 804-811.

[3] Gopal, R., Marsden, J. R., \& Vanthienen, J. (2011). Information mining—Reflections on recent advancements and the road ahead in data, text, and media mining. Decision support systems, 51(4), 727-731.

[4] Mendoza, M. L. Z., Agarwal, S., Blackshaw, J. A., Bol, V., Fazzi, A., Fiorini, F., ... \& Smith, A. M. (2024). A framework for developing a knowledge management platform. arXiv preprint arXiv:2406.12313.

[5] Chalkidis, I., Fergadiotis, M., Manginas, N., Katakalou, E., \& Malakasiotis, P. (2021). Regulatory compliance through Doc2Doc information retrieval: A case study in EU/UK legislation where text similarity has limitations. arXiv preprint arXiv:2101.10726.

[6] Small, S. G., \& Medsker, L. (2014). Review of information extraction technologies and applications. Neural computing and applications, 25, 533-548.

[7] Shu, D., Zhao, H., Liu, X., Demeter, D., Du, M., \& Zhang, Y. (2024, October). LawLLM: Law large language model for the US legal system. In Proceedings of the 33rd ACM International Conference on Information and Knowledge Management (pp. 4882-4889).

[8] Ke, Y. H., Jin, L., Elangovan, K., Abdullah, H. R., Liu, N., Sia, A. T. H., ... \& Ting, D. S. W. (2025). Retrieval augmented generation for 10 large language models and its generalizability in assessing medical fitness. npj Digital Medicine, 8(1), 187.

[9] Edge, D., Trinh, H., Cheng, N., Bradley, J., Chao, A., Mody, A., ... \& Larson, J. (2024). From local to global: A graph rag approach to query-focused summarization. arXiv preprint arXiv:2404.16130.

[10] Xia, Y., Zhou, J., Shi, Z., Chen, J., \& Huang, H. (2025, April). Improving retrieval augmented language model with self-reasoning. In Proceedings of the AAAI conference on artificial intelligence (Vol. 39, No. 24, pp. 25534-25542).

[11] Gao, Y., Xiong, Y., Gao, X., Jia, K., Pan, J., Bi, Y., ... \& Wang, H. (2023). Retrieval-augmented generation for large language models: A survey. arXiv preprint arXiv:2312.10997, 2, 1.

[12] Nauta, M., Trienes, J., Pathak, S., Nguyen, E., Peters, M., Schmitt, Y., ... \& Seifert, C. (2023). From anecdotal evidence to quantitative evaluation methods: A systematic review on evaluating explainable ai. ACM Computing Surveys, 55(13s), 1-42.

[13] Rave, J. I. P., Álvarez, G. P. J., \& Morales, J. C. C. (2021). Multi-criteria decision-making leveraged by text analytics and interviews with strategists. Journal of Marketing Analytics, 10(1), 30.

[14] Wu, Y., Zhang, Z., Kou, G., Zhang, H., Chao, X., Li, C. C., ... \& Herrera, F. (2021). Distributed linguistic representations in decision making: Taxonomy, key elements and applications, and challenges in data science and explainable artificial intelligence. Information Fusion, 65, 165-178.

[15] Ghisellini, R., Pareschi, R., Pedroni, M., \& Raggi, G. B. (2025). Recommending Actionable Strategies: A Semantic Approach to Integrating Analytical Frameworks with Decision Heuristics. arXiv preprint arXiv:2501.14634.

[16] Vaidya, O. S., \& Kumar, S. (2006). Analytic hierarchy process: An overview of applications. European Journal of operational research, 169(1), 1-29.

[17] Saaty, R. W. (1987). The analytic hierarchy process—what it is and how it is used. Mathematical modelling, 9(3-5), 161-176.

[18] Attri, R., Dev, N., \& Sharma, V. (2013). Interpretive structural modelling (ISM) approach: an overview. Research journal of management sciences, 2319(2), 1171.

[19] Singh, R., \& Bhanot, N. (2020). An integrated DEMATEL-MMDE-ISM based approach for analysing the barriers of IoT implementation in the manufacturing industry. International Journal of Production Research, 58(8), 2454-2476.

[20] Lei, H., Sun, C., Nie, M., Chen, X., Dong, Q., \& Ma, F. (2024, May). Using the AHP-TOPSIS Integrated Model to Assess the Quality of Urban Environments. In International Conference on Artificial Intelligence for Society (pp. 3-15). Cham: Springer Nature Switzerland.

[21] Blei, D. M., Ng, A. Y., \& Jordan, M. I. (2003). Latent dirichlet allocation. Journal of machine Learning research, 3(Jan), 993-1022.

[22] Mikolov, T., Chen, K., Corrado, G., \& Dean, J. (2013). Efficient estimation of word representations in vector space. arXiv preprint arXiv:1301.3781.

[23] Zhang, J., Gui, W., \& Wen, J. (2024). China’s policy similarity evaluation using LDA model: An experimental analysis in Hebei province. Journal of Information Science, 50(2), 515-530.

[24] Granger, C. W. (1969). Investigating causal relations by econometric models and cross-spectral methods. Econometrica: journal of the Econometric Society, 424-438.

[25] Spirtes, P., Glymour, C. N., \& Scheines, R. (2000). Causation, prediction, and search. MIT press.

[26] Drury, B., Oliveira, H. G., \& de Andrade Lopes, A. (2022). A survey of the extraction and applications of causal relations. Natural Language Engineering, 28(3), 361-400.

[27] Wang, J., Fu, J., Wang, R., Song, L., \& Bian, J. (2025). PIKE-RAG: sPecIalized KnowledgE and Rationale Augmented Generation. arXiv preprint arXiv:2501.11551.

[28] Liang, L., Sun, M., Gui, Z., Zhu, Z., Jiang, Z., Zhong, L., ... \& Zhou, J. (2024). Kag: Boosting llms in professional domains via knowledge augmented generation. arXiv preprint arXiv:2409.13731.

[29] Sarawagi, S. (2008). Information extraction. Foundations and Trends® in Databases, 1(3), 261-377.

[30] Sheikh, M., \& Conlon, S. (2012). A rule-based system to extract financial information. Journal of Computer Information Systems, 52(4), 10-19.

[31] Vásquez, J. A., Escobar, J. W., \& Manotas, D. F. (2021). AHP–TOPSIS methodology for stock portfolio investments. Risks, 10(1), 4.

[32] Brown, M., \& Shackel, P. (2023). Text mining oral histories in historical archaeology. International Journal of Historical Archaeology, 27(3), 865-881.

[33] O’Connor, B., Bamman, D., \& Smith, N. A. (2011, December). Computational text analysis for social science: Model assumptions and complexity. In Second workshop on comptuational social science and the wisdom of crowds (NIPS 2011).

[34] Zhang, Z., You, W., Wu, T., Wang, X., Li, J., \& Zhang, M. (2025, January). A Survey of Generative Information Extraction. In Proceedings of the 31st International Conference on Computational Linguistics (pp. 4840-4870).

[35] Dagdelen, J., Dunn, A., Lee, S., Walker, N., Rosen, A. S., Ceder, G., ... \& Jain, A. (2024). Structured information extraction from scientific text with large language models. Nature Communications, 15(1), 1418.

[36] Lewis, P., Perez, E., Piktus, A., Petroni, F., Karpukhin, V., Goyal, N., ... \& Kiela, D. (2020). Retrieval-augmented generation for knowledge-intensive nlp tasks. Advances in neural information processing systems, 33, 9459-9474.

[37] Zhu, Y., Wang, X., Chen, J., Qiao, S., Ou, Y., Yao, Y., ... \& Zhang, N. (2024). Llms for knowledge graph construction and reasoning: Recent capabilities and future opportunities. World Wide Web, 27(5), 58.

[38] Yang, Y., Peng, Q., Wang, J., \& Zhang, W. (2024). Multi-llm-agent systems: Techniques and business perspectives. arXiv preprint arXiv:2411.14033.

[39] Zangari, A., Marcuzzo, M., Rizzo, M., Giudice, L., Albarelli, A., \& Gasparetto, A. (2024). Hierarchical text classification and its foundations: A review of current research. Electronics, 13(7), 1199.

[40] Günther, M., Mohr, I., Williams, D. J., Wang, B., \& Xiao, H. (2024). Late chunking: contextual chunk embeddings using long-context embedding models. arXiv preprint arXiv:2409.04701.

[41] Ahmed, V., Khatri, M. F., Bahroun, Z., \& Basheer, N. (2023). Optimizing smart campus solutions: An evidential reasoning decision support tool. Smart Cities, 6(5), 2308-2346.

[42] Kotonya, N., \& Toni, F. (2024). Towards a framework for evaluating explanations in automated fact verification. arXiv preprint arXiv:2403.20322.

[43] Prasad, A., Saha, S., Zhou, X., \& Bansal, M. (2023). Receval: Evaluating reasoning chains via correctness and informativeness. arXiv preprint arXiv:2304.10703.

[44] Eberhard, K. (2023). The effects of visualization on judgment and decision-making: a systematic literature review. Management Review Quarterly, 73(1), 167-214.

[45] Oral, E., Chawla, R., Wijkstra, M., Mahyar, N., \& Dimara, E. (2023). From information to choice: A critical inquiry into visualization tools for decision making. IEEE Transactions on Visualization and Computer Graphics, 30(1), 359-369.
}

\newpage

\appendix

\section{Additional Experiments}

The RAD method supports complex decision-making by extracting criteria from domain documents and constructing a hierarchy with weight assignments. Therefore, the quality of the hierarchical model directly affects the credibility and transparency of the decision. This experiment aims to evaluate the rationality of the hierarchical decision model constructed by RAD in the following three key aspects:

\begin{itemize}
    \item Applicability of criteria: whether the extracted criteria truly reflect user requirements.
    \item Correctness of hierarchy: whether the hierarchical relationships among criteria are accurately constructed.
    \item Reasonableness of weight distribution: whether the relative importance aligns with the user's context.
\end{itemize}

To achieve this, we designed three specific experimental tasks. In the absence of a "gold standard" or expert judgment, we transformed the evaluation into explicit choice tasks and let the large model make the decisions. This approach helps reduce the risk of subjective hallucinations and allows for a quantitative assessment of accuracy.

\subsection{Experimental Setup}

We collected four documents each from the Federal Register (https://www.federalregister.gov/) related to the electric vehicle industry and wildlife conservation. These rules and notices have clear structural relationships, which are suitable for testing the hierarchical model. The datasets used are as follows:

\begin{enumerate}
    \item Electric Vehicle Industry Documents (EVID): Multi-Pollutant Emissions Standards for Model Years 2027 and Later Light-Duty and Medium-Duty Vehicles; National Electric Vehicle Infrastructure Standards and Requirements; Securing the Information and Communications Technology and Services Supply Chain Connected Vehicles; Statutory Updates to the Advanced Technology Vehicles Manufacturing Program.
    \item Endangered Species Protection Documents (ESPD): Designation of Experimental Populations; Enhancement of Survival and Incidental Take Permits; Regulations for Interagency Cooperation; Regulations Pertaining to Endangered and Threatened Wildlife and Plants.
\end{enumerate}

For each dataset, we constructed 15 simulated user decision requirements (30 in total). These cover various typical scenarios such as technical assessment, policy risk control, and policy formulation strategies. For each requirement, we used the RAD method to automatically build a weighted hierarchical decision model with 10 criteria. DeepSeek-V3 was used as the evaluation model. The evaluation approach transformed complex judgment tasks into multiple-choice questions with clear options and brief explanations.

\subsection{Experiment 3: Evaluation of Criterion Applicability}

\subsubsection{Experimental process}
In the criterion applicability evaluation, we first constructed 15 user decision requirements for each dataset. From the hierarchical decision model built by RAD, we randomly selected one criterion as a candidate. At the same time, we constructed a distractor criterion from a semantically unrelated text segment in the dataset as a comparison.

We then asked the LLM a closed-ended question: “For this user requirement, which criterion is more suitable to be included in the decision model?” The model was required to choose between the two criteria and provide a brief explanation.

Each user requirement was tested in 20 rounds, producing a total of $15*20$ judgment results. We measured the model’s ability to judge the semantic match between the criteria and user requirements by calculating the proportion of times it selected the correct criterion.

\subsubsection{Experimental Results}

The experimental results are shown in Tab. \ref{tab:ex3}.

\begin{table}
    \caption{The results of Experiment 3.}
    \centering
    \label{tab:ex3}
    \begin{tabular}{ccc}
    \toprule
        Dataset & Proportion of RAD Selected & Binomial Test p-value (RAD > 0.5) \\
        EVID & 0.9467 &  p < 0.001 \\
        ESPD & 0.9567 &  p < 0.001 \\
    \bottomrule
    \end{tabular}
\end{table}

\subsubsection{Analysis}
Experiment 3 aims to evaluate the ability of the RAD method to identify appropriate criteria in the "criterion applicability" task. The results show that the proportion of RAD criteria selected reached 94.67\% for the electric vehicle industry dataset, and 95.67\% for the endangered species protection dataset. One-sided binomial tests (assuming a random selection probability of 50\%) were conducted for both datasets, with p-values significantly below 0.001, indicating that the criteria selected by RAD are significantly better than random.

These results suggest that the criteria extracted by RAD from domain documents are highly aligned with actual user requirements. The LLM achieved an accuracy of around or above 95\% in the binary forced-choice task, demonstrating the effectiveness of the criterion extraction module. The small differences between domains may reflect variations in language style across user requirements, but overall performance remains robust.

\subsection{Experiment 4: Evaluation of Hierarchical Structure}

\subsubsection{Experimental Process}

This experiment focuses on evaluating the hierarchical structure built by the RAD method, specifically whether the parent-child relationships between nodes in the multi-level decision tree are reasonable. For each user requirement, we randomly select two criterion nodes with adjacent hierarchical levels from the RAD model—for example, a primary indicator and its subordinate sub-indicator. These two nodes are then presented to the LLM, along with the question of whether a hierarchical relationship exists between them, and if so, which should be the parent node. The model must choose from three options: "correct parent-child relationship," "reversed relationship," or "no hierarchical relationship."

Each user requirement is used to generate 20 pairs of criteria, resulting in a total of $15*20$ sets of judgments. We then compare the model’s responses with the original hierarchy constructed by the RAD method to assess the accuracy and consistency of its semantic structure understanding.

\subsubsection{Experimental Results}

The experimental results are shown in Tab. \ref{tab:ex4}.

\begin{table}
    \caption{The results of Experiment 4.}
    \centering
    \label{tab:ex4}
    \begin{tabular}{cccc}
    \toprule
        Dataset & Proportion of RAD Selected & Binomial Test p-value (RAD > 0.5) \\
        EVID & 0.9600 &  p < 0.001 \\
        ESPD & 0.9267 &  p < 0.001 \\
        \bottomrule
    \end{tabular}
\end{table}

\subsubsection{Analysis}

The second experiment aims to verify whether the parent-child relationships in the hierarchical model built by RAD align with semantic commonsense and decision-making requirements. The results show that the consistency between the judgments and the original RAD structure reaches 96.00\% in the electric vehicle industry dataset and 92.67\% in the endangered species protection dataset. One-sided binomial tests for both datasets show a statistically significant advantage over random selection (p < 0.001), indicating that the model is highly effective in identifying hierarchical relationships.

The slightly better performance in the electric vehicle domain may be due to the clearer technical terms and classification logic. In contrast, the wildlife protection domain presents more semantic overlap between concepts, slightly increasing the difficulty of judgment. Overall, the hierarchical structures built by the RAD method demonstrate high semantic validity and stability, and maintain consistency across different domains when repeatedly tested by the LLM through closed-form questions.

\subsection{Experiment 5: Evaluation of Weight Appropriateness}

\subsubsection{Experimental Process}

In this experiment, we assess the reasonableness of weight assignments in the RAD-generated hierarchy. We randomly select a set of sub-criteria at the same level from the hierarchical model, along with their normalized weights, as the original output of the RAD method. Then, we manually construct an alternative weight scheme, which maintains the same total sum but adjusts the distribution of weights among some criteria.

Both the original and the alternative weight schemes are presented to the LLM, along with the corresponding user decision context. The model is asked: “Which set of weights better reflects the relative importance of these criteria for this decision-making requirement?” The model must choose one of the two options and briefly explain the reason.

Each decision requirement is evaluated through 20 rounds of different weight judgment experiments, resulting in a total of $15*20$ evaluation results. The proportion of times the RAD original weight scheme is preferred by the model serves as the evaluation metric, reflecting the semantic appropriateness of RAD's weight assignment.

\subsubsection{Experimental Results}

The experimental results are shown in Tab. \ref{tab:ex5}.

\begin{table}
    \caption{The results of Experiment 5.}
    \centering
    \label{tab:ex5}
    \begin{tabular}{cccc}
    \toprule
        Dataset & Proportion of RAD Selected & Binomial Test p-value (RAD > 0.5) \\
        EVID & 0.6733 & p < 0.001  \\
        ESPD & 0.7366 & p < 0.001  \\
        \bottomrule
    \end{tabular}
\end{table}

\subsubsection{Analysis}

Experiment 3 aims to evaluate the semantic appropriateness of RAD’s weight assignment, specifically by testing its performance in judging weight preferences among criteria at the same hierarchical level. As shown in the table, the original RAD weight scheme was selected in 67.33\% of cases for the electric vehicle dataset and 73.66\% for the endangered species dataset. Both results are significantly above the random baseline (p < 0.001), indicating that the large model generally prefers the original normalized weight scheme.

However, compared with the previous two experiments, the accuracy here is noticeably lower. This may be due to the inherently subjective nature of weight assignment. Different evaluators (or models) may have varying perceptions of the importance of different criteria. Moreover, subtle numerical differences between weight schemes are not easily detected in a closed-choice format. When the semantic boundaries of criteria are unclear, the model tends to make random choices or rely on additional contextual cues introduced in the prompt.

\end{document}